\pdfoutput=1

\documentclass[11pt]{article}

\usepackage[final]{acl}

\usepackage{times}
\usepackage{latexsym}

\usepackage[T1]{fontenc}

\usepackage[utf8]{inputenc}

\usepackage{microtype}

\usepackage{inconsolata}

\usepackage{booktabs}
\usepackage{graphicx}
\usepackage{colortbl}
\usepackage{caption}
\usepackage{subcaption}
\usepackage{comment}
\usepackage{tikz}

\title{Modeling and Detecting Company Risks from News}

\author{
    Jiaxin Pei$^{\dagger}$\thanks{~~Work done as an intern at Bloomberg.} ~ Soumya Vadlamannati$^{\ddagger}$ ~ Liang-Kang Huang$^{\ddagger}$ \\
    {\bf Daniel Preo\c{t}iuc-Pietro$^{\ddagger}$} ~ {\bf Xinyu Hua$^{\ddagger}$}\\
$^\dagger$University of Michigan, Ann Arbor, MI, USA  \\
$^\ddagger$Bloomberg, New York, NY, USA\\
{\tt pedropei@umich.edu}\\
{\tt  \{svadlamanna1, lhuang214, dpreotiucpie, xhua22\}@bloomberg.net}
}

\begin{document}
\maketitle
\begin{abstract}
Identifying risks associated with a company is important to investors and the wellbeing of the overall financial markets. In this study, we build a computational framework to automatically extract company risk factors from news articles. 
Our newly proposed schema comprises seven distinct aspects, such as supply chain, regulations, and competition.
We annotate $666$ news articles and benchmark various machine learning models.
While large language models have achieved remarkable progress in various types of NLP tasks, our experiment shows that zero-shot and few-shot prompting state-of-the-art LLMs (e.g., Llama-2) can only achieve moderate to low performances in identifying risk factors. In contrast, fine-tuning pre-trained language models yields better results on most risk factors. Using this model, we analyze over 277K Bloomberg News articles and demonstrate that identifying risk factors from news could provide extensive insights into the operations of companies and industries. 
\end{abstract}

\section{Introduction}
Risks are inherent and pervasive within companies' operations and our society \citep{stephany2022corisk, rausand2013risk, albuquerque2019corporate}. Understanding and identifying corporate risk factors could benefit diverse stakeholders, including investors, regulators, and other relevant entities. Notably, publicly listed companies are mandated to disclose their risk factors, as these can inform shareholders and the public when making financial decisions \citep{beretta2004framework}. NLP models are also built to automatically extract company-related risk factors from public filings, providing consolidated and accessible insights for analysts to fathom and integrate these factors \citep{kogan2009predicting, yang2018corporate}.

While company filings offer a systematic view of company-related risks, they are beset by three principal issues: (1) Limited frequency—owing to mandatory quarterly reporting, risk analysis is confined to three-month intervals, disregarding the reality of swift, even daily, alterations in a company's risk profile. (2) Subjectivity—authored by internal personnel, company filings might inadvertently omit pivotal risk factors due to vested interests \citep{masson2020detecting, klingebiel2018risk}. (3) Bias towards public entities—only publicly listed companies are obligated to divulge risks through filings, neglecting the imperative to comprehend risks associated with private companies, which may be particularly pertinent when engaged in financial activities such as bond issuance~\citep{abdel1993private, vanstraelen2017auditing}. 

To redress these limitations, we propose to model company-related risk factors from news articles. News articles offer the following merits in analyzing company-related risk factors: (1) High frequency—news updates occur in real time, providing a dynamic information stream conducive to measuring companies' risk factors in the ever-evolving market. (2) External perspective—news articles, devoid of company affiliations, proffer diverse viewpoints, shedding light on risk factors from external vantage points.  (3) Coverage over both public and private companies—news articles encapsulate both publicly listed and private companies, thereby bridging the information gap present in public filings.

While there are existing studies on modeling risk factors, they typically focus on company filings \citep{zhu2016firm, kravet2013textual} and their categorization may not be directly applicable to news data. 
Combining existing literature and our manual examination of hundreds of news articles, we propose a novel theoretical framework to analyze company risk factors in news. Our taxonomy encompasses seven categories of risk factors: 
\texttt{Supply Chain and Product}, \texttt{People and Management}, \texttt{Finance}, \texttt{Legal and Regulations}, \texttt{Macro}, \texttt{Competition}, and \texttt{Markets and Consumers}.

We annotate $666$ news articles from Bloomberg News\footnote{\url{https://www.bloomberg.com}} and benchmark a series of models from feature-engineered baselines to prompting large language models (LLMs). 
Despite their impressive results on various other NLP tasks, LLMs perform worse than smaller transformer models (e.g., RoBERTa) fine-tuned on in-domain data.
By applying the best-performing model to a large sample of $277,112$ news articles, we analyze the risk factor across companies in given industries, and across the entire macroeconomy.  
Our analysis shows that modeling company-related risk factors could reveal important signals of not only companies' operations but also can be used as the indicator of the macro-level risk for society.

\section{Related Work}

\subsection{Risk Factors in Finance Domain}
\label{subsec:risk_in_finance}
In finance and corporate operations, risks refer to the factors that may harm the company or may cause it to fail \citep{rausand2013risk}. Existing research has identified many types of risk factors for companies, including financial risk \citep{malz2011financial, fujii2022extraction}, credit risk \citep{kao2000estimating}, policy risk \citep{blyth2007investment}, macro risk \citep{hiang2006macroeconomic}, operational risk \cite{fujii2022extraction} and competition risk \citep{raith2003competition}. In many countries, regulators require publicly listed companies to disclose risk factors in their quarterly and annual reports to inform the investors~\citep{weil2006effectiveness}. Another line of research relevant to risk factors is financial and economic uncertainty~\citep{moore2017measuring}. 
While both are forward-looking, risks specifically connotate those factors that may negatively affect the operations and market value of a company.

\subsection{Risk Identification as an NLP Task}
Natural Language Processing methods have long been used to analyze text documents in the finance domain~\citep{loughran2020textual}, such as news reports~\citep{day2016deep}, social media posts~\citep{souza2015twitter}, and company filings~\citep{wang2013financial}. Existing studies primarily focus on company filings~\citep{wang2013financial, yang2018corporate, kogan2009predicting}, which are issued by companies themselves and are subject to limited frequency.
While researchers have also extracted risk factors from news articles~\citep{lu2009identifying, bhadani2020mining}, these either focus only on specific types of risks~\citep{bhadani2020mining} or only tried to identify the relevant claims~\citep{lu2009identifying} instead of providing a holistic view of the risks.

\section{Theoretical Framework}
\label{sec:theory}
In this study, we focus on modeling company-level risk factors in news articles. 
We survey existing literature (\S~\ref{subsec:risk_in_finance}) and qualitatively examine hundreds of news articles.
Our proposed framework is summarized as below.

\paragraph{Supply Chain and Product} Risks associated with the company's supply chain, manufacturing, product or core technology. For example, ``\textit{Yum China Faces Challenges with Chicken Prices}'' indicates risks regarding the supply chain, as chicken is an important ingredient in Yum China’s products.

\paragraph{People and Management} Risks regarding a company's internal operations such as layoffs, departures of top management, or specific operation strategies. For example, ``\textit{Tesla Pauses Hiring, Musk Says Need to Cut Staff by 10\%}.''

\paragraph{Finance} Risks related to the finances of a company such as cash flow, fund procurement, investments, and profits. For example, ``\textit{NIO Shares Soar as Loss Shrinks, Though Cash Concerns Linger}.''

\paragraph{Legal and Regulations} Risks induced by potential policy changes, pressure from regulations or lawsuits. For example, ``\textit{Maple Leaf Plunges as China's Hog Suspension Impacts Profits}.''

\paragraph{Macro} Risks caused by the macro socio-economic environment such as inflation, pandemics or a financial crisis. For example, ``\textit{Absa Drops on Profit Miss as South African Economy Struggles}.'' 

\paragraph{Markets and Consumers} Risks or challenges from the market or consumer sales. For example, ``\textit{Hong Kong Protests Cut Demand for Hilton, Hyatt Hotel Rooms}'' suggests that the demand for hotel space is shrinking, which indicates \texttt{Markets and Consumer} risk for both Hilton and Hyatt.

\paragraph{Competition} Risks from a company's competitors in the market. For example, ``\textit{Apple Revamping Smart-Home Efforts to Challenge Amazon, Google}.''

\begin{figure*}[t]
\centering
\includegraphics[width=160mm]{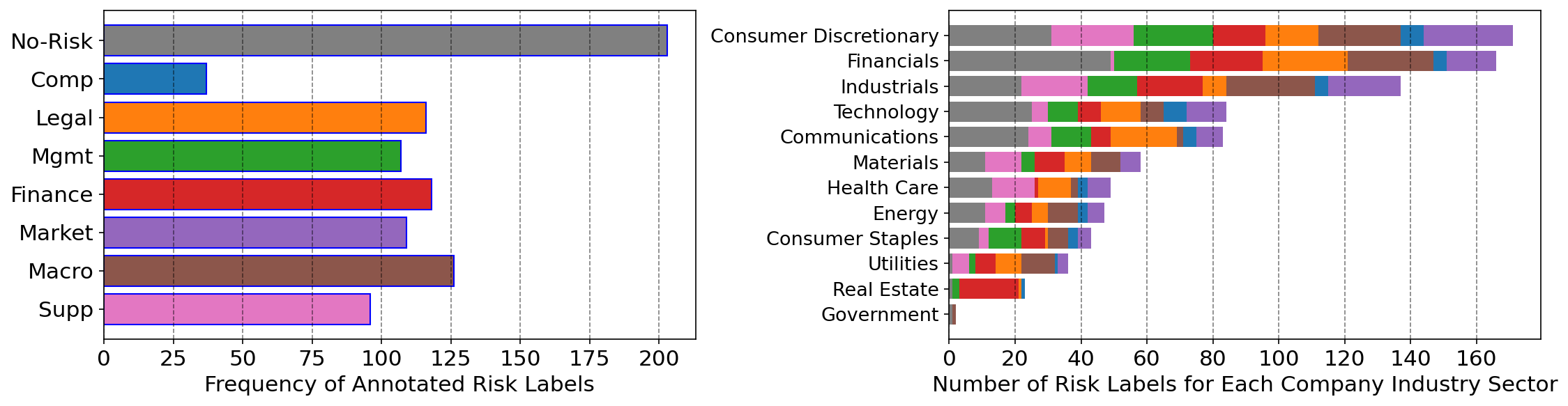}

\caption{
[Left] Label distribution over the 716 annotated samples. We denote \texttt{Competition} as \texttt{Comp}, \texttt{Legal and Regulations} as \texttt{Legal}, \texttt{People and Management} as \texttt{Mgmt}, and \texttt{Supply Chain and Product} as \texttt{Supp}. All except \texttt{Competition} have an approximately $14\%$ positive rate. In total, $71.6\%$ of the samples are labeled with at least one risk factor. [Right] Distribution of risk labels for each company industry sector, based on the Bloomberg Industry Classification Standard (BICS). In total, 12 different sectors are covered in the annotated dataset.
}
\label{fig:label-dist}
\end{figure*}

\section{Data and Annotation}
\label{sec:data}

\paragraph{BN Dataset} 
We draw five years of articles published by Bloomberg News\footnote{\url{https://www.bloomberg.com}} (hereafter \textbf{BN}), covering diverse events and opinions pertaining to companies across the world. We choose news published between 2018 and 2022 to allow for COVID-related comparisons. 
This initial dataset is filtered by removing machine-generated and non-natural language content, and is preprocessed with a rule-based entity extraction pipeline. We further removed articles where no company is mentioned. 
This results in a collection of $277,112$ news articles covering $14,972$ public and $11,413$ private companies. For the sake of simplicity, we keep only the headline and first five sentences of each article for our study. 
These articles range from $20$ to $4,430$ tokens long, with the average article being $151$ tokens.

\begin{table}[h]
\fontsize{9}{11}\selectfont
\centering
\setlength{\tabcolsep}{1mm}
\begin{tabular}{l cccccc}
    \toprule
    Year & 2018 & 2019 & 2020 & 2021 & 2022 & Total \\
    \midrule
    \# Docs. & 56,741 & 62,862 & 57,421 & 53,906 & 46,182 & 277,112  \\
    \bottomrule
\end{tabular}

\end{table}

\paragraph{Risk-Related Pre-Filtering}
Our pilot study shows that risk factors can be sparse in news.
Directly annotating over a random subset of BN articles will therefore yield a very high negative ratio. 
We apply a lexicon-based filter before the sampling. Concretely, we iteratively curate 53 unigrams to capture various aspects of risk events, such as ``\textit{challenge},'' ``\textit{layoff},'' ``\textit{shrink}.''\footnote{Full list can be found in Appendix.} We consider annotating an article only if its headline matches at least one of the keywords. We also experimented with hedges similar to~\citep{pei2021measuring}. However, we found that while hedges are a good proxy for uncertainty, they are not able to reliably recall news articles regarding risks, highlighting the difference between uncertainty and risk factors.

\paragraph{Annotation} 
We conduct an annotation study based on the seven risk factors mentioned in Section \ref{sec:theory}, using a multi-label classification setting.
We hire three U.S.-based annotators who are experienced in the finance domain. They are first instructed to label $100$ articles independently, followed by a discussion to resolve disagreements and make modifications to the annotation guideline.
The adjudicated set is used as test data. 

They further annotate $200$ articles each. After removing samples with wrong mentions or low-quality text, the final dataset includes $716$ samples from $666$ unique news articles. In this dataset, $49\%$ of the samples have exactly one label, while more than $20\%$ mention multiple factors~\footnote{Detailed distribution is in Appendix (\S~\ref{sec:label-count})}. 
In Figure~\ref{fig:label-dist} we show the distribution of risk factors along with the number of news without any risk factors (``\textsc{No-risk}'').

\begin{figure}[h]
\fontsize{9}{9.5}\selectfont
    \centering
     \setlength{\tabcolsep}{0.5mm}
    \begin{tabular}{|l|ccccccc|}
    \hline
      & \textsc{Supp} & \textsc{Mgmt} &  \textsc{Fin} & \textsc{Legal} & \textsc{Macro} & \textsc{Comp} & \textsc{Mrkt} \\
      \hline

\textsc{ Supp } & \cellcolor{red!0}N/A & \cellcolor{red!8}8 & \cellcolor{red!6}6 & \cellcolor{red!12}12 & \cellcolor{red!17}17 & \cellcolor{red!8}8 & \cellcolor{red!19}19   \\
\textsc{ Mgmt } & \cellcolor{red!8}8 & \cellcolor{red!0}N/A & \cellcolor{red!15}15 & \cellcolor{red!16}16 & \cellcolor{red!12}12 & \cellcolor{red!1}1 & \cellcolor{red!12}12   \\
\textsc{ Fin } & \cellcolor{red!6}6 & \cellcolor{red!15}15 & \cellcolor{red!0}N/A & \cellcolor{red!8}8 & \cellcolor{red!22}22 & \cellcolor{red!0}0 & \cellcolor{red!9}9   \\
\textsc{ Legal } & \cellcolor{red!12}12 & \cellcolor{red!16}16 & \cellcolor{red!8}8 & \cellcolor{red!0}N/A & \cellcolor{red!11}11 & \cellcolor{red!4}4 & \cellcolor{red!8}8   \\
\textsc{ Macro } & \cellcolor{red!17}17 & \cellcolor{red!12}12 & \cellcolor{red!22}22 & \cellcolor{red!11}11 & \cellcolor{red!0}N/A & \cellcolor{red!3}3 & \cellcolor{red!31}31   \\
\textsc{ Comp } & \cellcolor{red!8}8 & \cellcolor{red!1}1 & \cellcolor{red!0}0 & \cellcolor{red!4}4 & \cellcolor{red!3}3 & \cellcolor{red!0}N/A & \cellcolor{red!11}11   \\
\textsc{ Mrkt } & \cellcolor{red!19}19 & \cellcolor{red!12}12 & \cellcolor{red!9}9 & \cellcolor{red!8}8 & \cellcolor{red!31}31 & \cellcolor{red!11}11 & \cellcolor{red!0}N/A   \\

   \hline
    \end{tabular}
    \caption{%
   Risk co-occurrence matrix (annotated dataset).
    }
    \label{fig:label-coocur}
\end{figure}

\paragraph{Statistics} 
To better understand the characteristics of our newly annotated dataset, we first show the distribution of industry sectors to which the detected companies belong. We match each company to one of the 12 high-level industry sectors defined by  the Bloomberg Industry Classification Standard (BICS)~\footnote{\url{https://tinyurl.com/3nnzr3p9}}. 
As shown on the right side of Figure~\ref{fig:label-dist}, our dataset contains samples over all of these sectors. 
The distribution of risk types differs across industry sectors. For instance, there are more \textsc{Market} related risks for companies in ``\textit{Consumer Discretionary},'' while more ``\textit{Legal}'' risks are mentioned for companies in the ``\textit{Financials}'' and ``\textit{Communications}'' industries. For ``\textit{Real Estate}'', the majority of the risks fall under \textsc{Finance}.
In Figure~\ref{fig:label-coocur} we further illustrate the co-occurrence of risk factors. Notably, we observe higher co-occurrence of (\textsc{Finance}, \textsc{Macro}) and (\textsc{Market} and \textsc{Macro}) pairs.

\section{Benchmark}
We formulate the risk prediction task as a multi-label classification problem: given a news article and a mentioned company, we aim to predict whether each of the seven risk factors is mentioned. We consider non-neural baseline models, fine-tuning pre-trained transformers, and large-language models (LLM) with in-context learning. We split the dataset into $484$ samples for training, $126$ for validation, and $106$ for testing.

\subsection{Models}

We first experiment with non-neural baseline models: (1) \textbf{Random}: for each risk factor, randomly assign a binary label with equal probabilities. (2) \textbf{Logistic Regression}: we calculate TF-IDF (up to bigrams) features and run logistic regression models for each risk factor. Similarly, (3) \textbf{Support Vector Machine (SVM)} models are trained using the same TF-IDF features and linear kernel. (4) We further implement $k$-nearest neighbor (\textbf{KNN}) models using document embeddings calculated from a fine-tuned RoBERTa model~\cite{liu2019roberta} with SimCSE~\cite{gao2021simcse} objective.

\paragraph{Pre-trained Transformers with Fine-tuning} 
We benchmark common pre-trained transformer models as sequence classification tasks under a supervised fine-tuning setting: (1) \textbf{BERT-large}~\cite{devlin2018bert}, (2) \textbf{RoBERTa-base} and \textbf{RoBERTa-large}~\cite{liu2019roberta}, (3) \textbf{RoBERTa-large-BB}: a RoBERTa model further pre-trained on $13$ years of Bloomberg News data. 

\begin{figure}[t]
\centering
\hspace{-5mm}
\includegraphics[width=80mm]{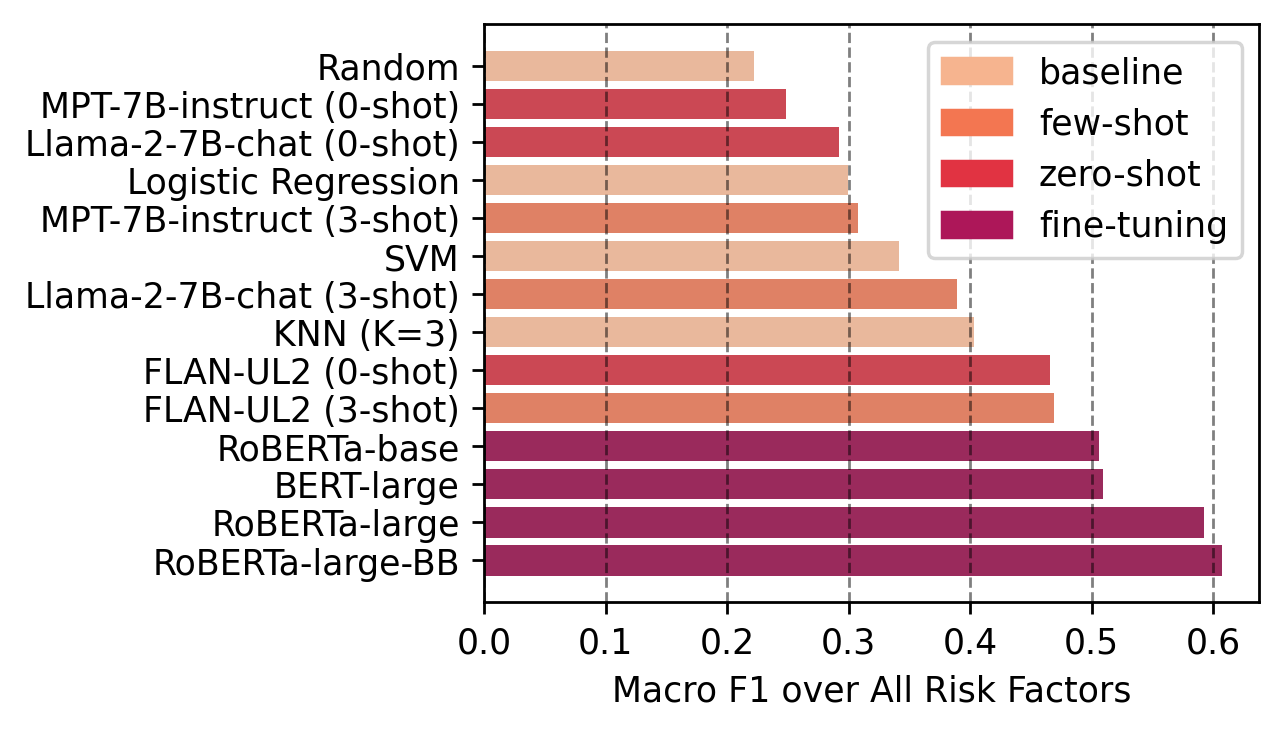}

\caption{
The overall performance of different models.
The best result is achieved by fine-tuning the \textbf{RoBERTa-large-BB} model, which is pre-trained on domain-specific datasets. Zero-shot and few-shot prompting for LLM perform worse than the fine-tuned models by a large margin.
}
\label{fig:model_performance}
\end{figure}

\begin{figure}[t]
\centering
\includegraphics[width=70mm]{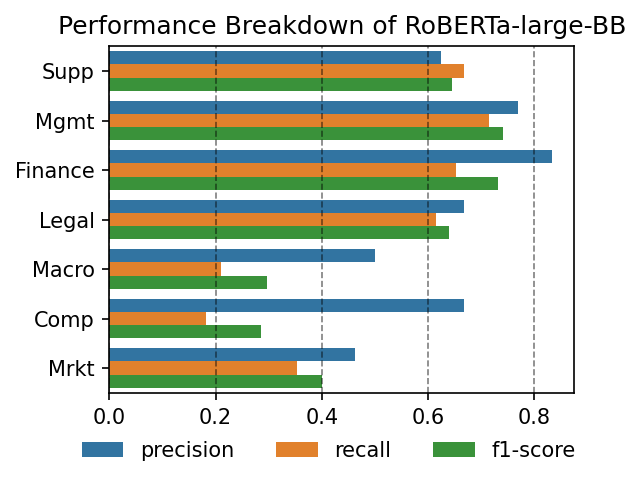}

\caption{
The best performance on each risk factor. Identifying company risks on \texttt{Macro}, \texttt{Markets and Consumer} and \texttt{Competition} remains hard.
}
\label{fig:performance_by_risk}
\end{figure}

\paragraph{LLM with Prompting} 
We further compare with three open source, instruction-tuned large language models (LLM) under the in-context learning setting: (1) \textbf{FLAN-UL2}\footnote{\url{https://www.yitay.net/blog/flan-ul2-20b}}, which is an instruction-tuned version of the UL2~\cite{tay2022ul2} model over the FLAN~\cite{longpre2023flan} dataset. (2) \textbf{MPT-7B-instruct}~\cite{MosaicML2023Introducing} is a decoder-only model with 7 billion parameters, trained on the \texttt{dolly-hhrlhf}~\footnote{\url{https://huggingface.co/datasets/mosaicml/dolly_hhrlhf}} dataset.
Lastly, (3) \textbf{Llama-2-7B-chat}~\cite{touvron2023llama} is a decoder-only model optimized for dialogue tasks, achieving competitive performance on various NLP tasks against closed-source LLMs. 

\begin{figure*}
    \centering
    \includegraphics[width=120mm]{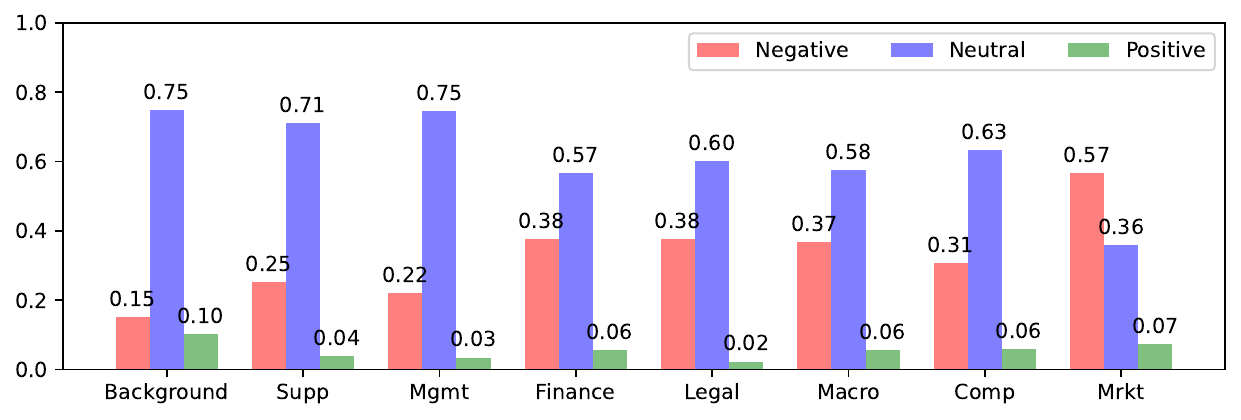}
    \caption{
While news articles mentioning risk factors are more negative overall compared with the overall distribution (background), both positive and neutral news can mention risk factors for companies.
}
\label{fig:sentiment_risk}
\end{figure*}

For each risk factor, we construct the following prompt template~\footnote{We empirically select prompt templates based on manual inspection of the performance.} with the input \texttt{news text}, the \texttt{target} company name, and a full description of the \texttt{risk} from the annotation guideline: 

\begin{table}[h]
\centering

\scalebox{0.95}{
\begin{tikzpicture}
\fontsize{9}{11}\selectfont
\node[draw, inner sep=5pt, rounded corners, fill=gray!20]{
    \begin{tabular}{p{65mm}}

    \texttt{\{news text\}} 
    
    \texttt{For company \{target\}, does the above news mention \{risk\} ?}
    
    \texttt{Options: Yes, No}
    
    \texttt{Your answer is (Please only use Yes or No):}
\end{tabular}
};
\end{tikzpicture}}

\label{tab:anno}
\vspace{-2mm}
\end{table}

We consider both the zero-shot and few-shot settings for all LLMs. The few-shot samples are chosen as the $k$-nearest neighbors ($k=3$) from the training set, which are represented using the same template and directly prepended to the test sample.

\subsection{Result}
Figure~\ref{fig:model_performance} shows the overall model performance. 
Fine-tuning transformers yields the best performance, especially the \textbf{RoBERTa-large-BB} model that is trained on a domain-specific dataset.
We breakdown the per-risk performance in Figure~\ref{fig:performance_by_risk}. The model achieves better results for \texttt{Supply Chain and Product} and \texttt{Finance}. However, identifying \texttt{Macro}, \texttt{Competition} and \texttt{Markets and Consumers} risks remain challenging.

\section{Application and Analysis}
Identifying company-related risk factors in news opens many potential applications. In this section, we explore the applications of our model over a large-scale Bloomberg News dataset.

\subsection{Are risk factors just negative sentiment?}
\label{sec:risk-and-sentiment}

\begin{figure*}
    \centering
    \includegraphics[width=140mm]{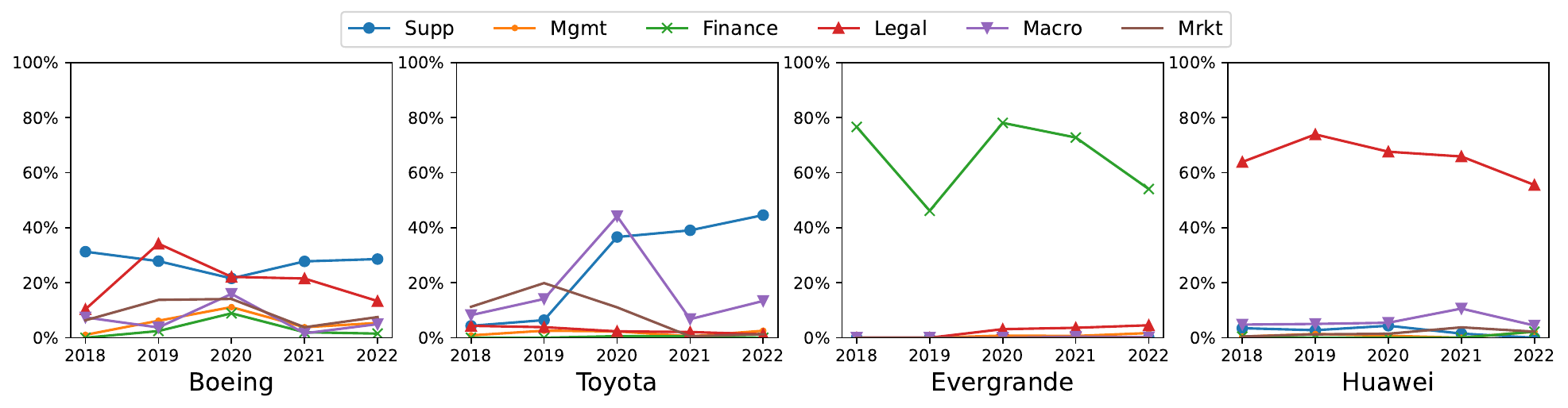}
    \caption{Percentage of news stories tagged by each risk factor type, for different companies. }
    \label{fig:risk_percentage_per_company}
\end{figure*}

The term ``risk'' inherently carries negative connotations. In practice, are risk factors simply negative sentiment? In this study, we explore the connection between company-level sentiment and risk factors. 
We run an off-the-shelf sentiment analysis model\footnote{This model is based on DistilBERT~\cite{Sanh2019DistilBERTAD} and is fine-tuned on finance news with sentiment labels.} over the large 5-year \textbf{BN} dataset. For each company mentioned in a news article, a probabilistic distribution over ``Positive,'' ``Neutral,'' and ``Negative'' is estimated. 

As shown in Figure~\ref{fig:sentiment_risk}, the overall sentiment for a company tends to be more negative when it faces risks. The largest gap of sentiment occurs for \texttt{Legal and Regulations} and \texttt{Markets and Consumers}, where risks are usually mentioned with negative sentiment. Nevertheless, risk factors can be mentioned even when the overall sentiment regarding a company is neutral or positive
(See Table~\ref{tab:sample-news-sentiment} in Appendix for examples). This suggests that risk factors are not just negative sentiment. 

\begin{figure*}[t]
\centering
\hspace{-5mm}
\includegraphics[width=145mm]{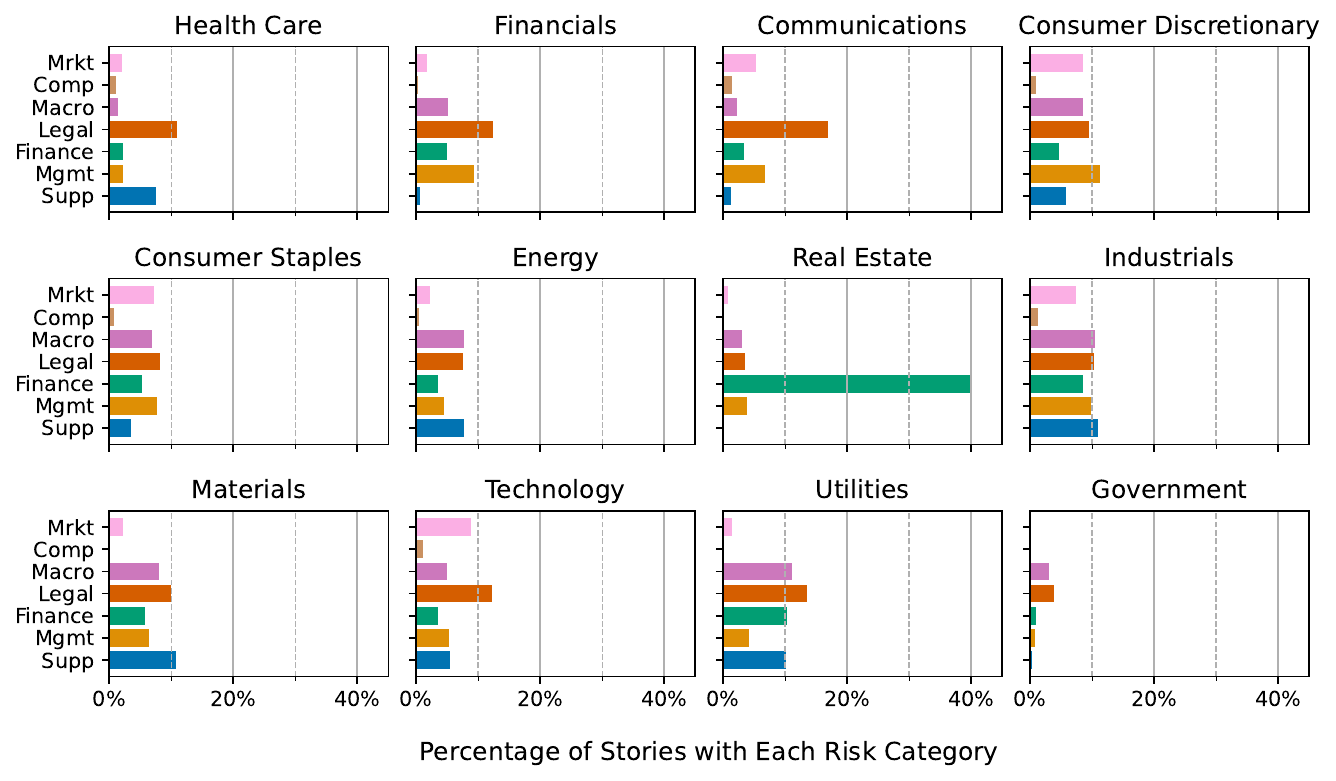}
\caption{
Risk distribution for companies in different industries.
}
\label{fig:industry_risk}
\end{figure*}

\begin{figure*}[h]
\centering
\includegraphics[width=130mm]{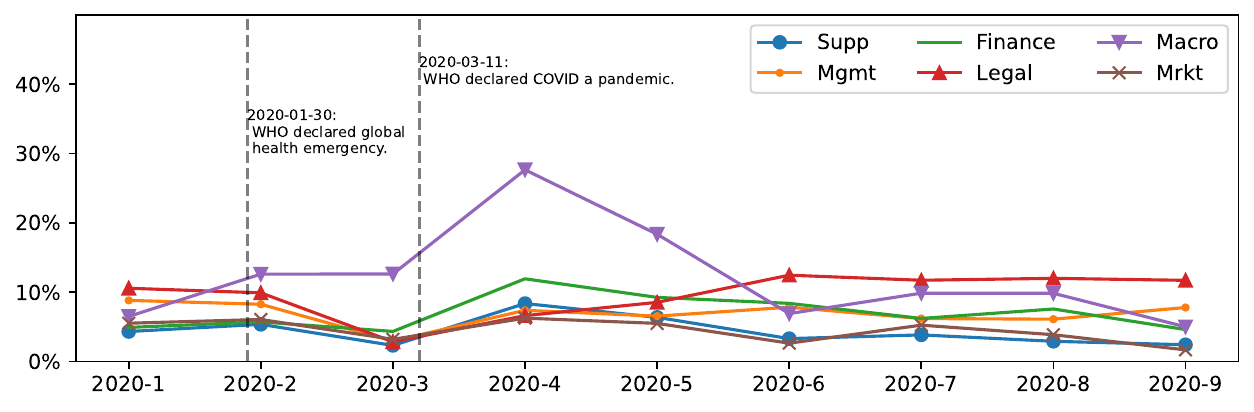}
\includegraphics[width=130mm]{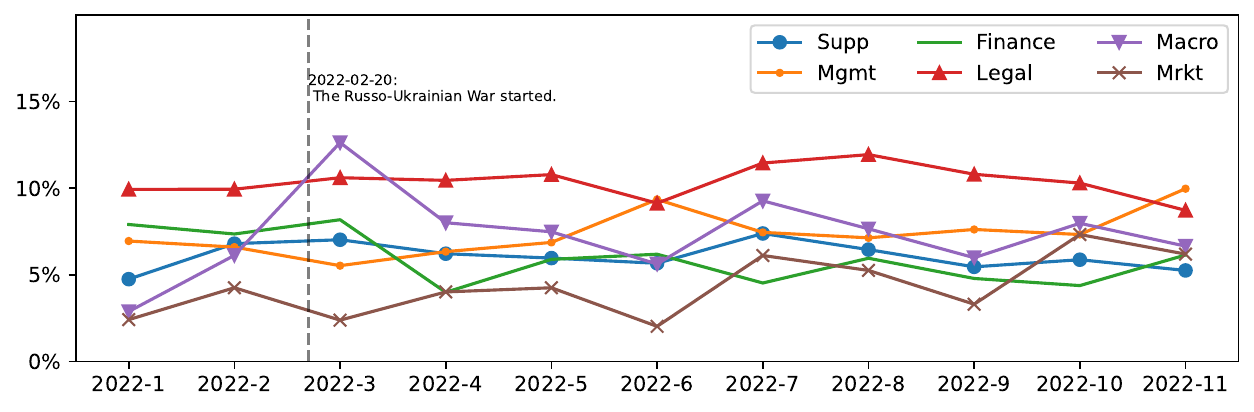}
\caption{
[Top] COVID-19 induces nearly all types of risk factors for companies.
[Bottom] Russia's invasion of Ukraine sees increased Macro risks.
}
\label{fig:monthly_risk_2020}
\end{figure*}

\subsection{Company-level Study}

\paragraph{Boeing} In 2018, the first 737 MAX airplane crashed into the Java sea. As shown in Figure~\ref{fig:risk_percentage_per_company}, Boeing faced high risks regarding its products (\texttt{Supp}) in 2018, while other types of risks generally remained low. In 2019, the second 737 MAX crashed, which immediately led to the involvement of the regulators (\texttt{Legal}). Risks related to the market, consumers, and management also rose in 2019.

\paragraph{Toyota} Motor Corporation is the largest carmaker in the world. From 2018 to 2020, Toyota saw major \texttt{Macro} and \texttt{Markets and Consumers} risk. In 2020, the world was faced with a global chip shortage, which further led to a spike in \texttt{Supply Chain and Product} risk for Toyota.

\paragraph{Evergrande} The Chinese real-estate company Evergrande Group has gone through various debt issues in recent years, which is reflected by the overwhelming percentage of \texttt{Financial} risks predicted by the model.

\paragraph{Huawei} Unlike company filings which only include publicly traded companies, news articles also allow us to analyze private companies. Huawei Technologies Co., Ltd. is the world's leading communication technology and phone producer. Since 2018, Huawei has faced regulatory risks from the U.S. government. Figure~\ref{fig:risk_percentage_per_company} shows the overall change in Huawei's risk factors mentioned in Bloomberg News data. Huawei saw major regulatory risks from 2018 to 2020. Because of these regulations, Huawei's market and sales are also affected, and it has seen higher \texttt{Macro} and \texttt{Market} risks since 2021.

\subsection{Industry-level Study}
Companies in different industries are different in nature and therefore may face different types of risk factors. In this section, we explore the risk factors associated with companies in different industries. We use Bloomberg internal company categorization and map each company to one of the 12 top industry categories: Health Care, Financials, Technology, Energy, Consumer Discretionary, Utilities, Communications, Real Estate, Consumer Staples, Industrials, Materials, and Government. The results are displayed in Figure~\ref{fig:industry_risk}.

\paragraph{Financials} Financial companies rarely see risks from the \texttt{Supply Chain and Product} side and are more likely to face risks from \texttt{People and Management} and \texttt{Legal and Regulations}. 

\paragraph{Real Estate} The real estate industry faced high \texttt{Finance} risk from 2018 to 2022, potentially due to the debt crisis of the real estate companies in China.

\paragraph{Health Care} The Health Care industries are associated with high \texttt{Legal and Regulation} and \texttt{Supply Chain and Product} risks, potentially due to the production of and regulations surrounding the COVID-19 vaccines, in addition to providing other health care services in response to the COVID-19 pandemic.

\paragraph{Others} Industries like Consumer Discretionary, Consumer Staples and Industrials generally see balanced risks across all factors.

\subsection{Macro-level Study}

\paragraph{COVID-19 Pandemic}

Since early 2020, the COVID-19 pandemic posed huge global challenges. Figure~\ref{fig:monthly_risk_2020} shows the aggregated risk factors in each month in 2020. The first COVID-19 case was identified in January 2020 and the World Health Organization (WHO) announced a global health emergency on January 31st in response to the rapid increase in infections and deaths worldwide. A new global health emergency led to a sharp rise of \texttt{Macro} risks in February. The world may still not have been fully aware of other types of risks, and therefore other risks remained stable in February. However, the situation was changing rapidly. In March, the WHO declared COVID-19 a pandemic and the United States officially issued a national emergency, which led to a sharp rise in all other risk factors in April.

\paragraph{Russia's Invasion of Ukraine} In February 2022, Russia invaded Ukraine and this event immediately led to an increase in \texttt{Macro} risks for companies. Similar to the beginning of COVID-19, other types of risks are not reflected at this early stage. However, in June, Russia cut natural gas supplies by more than half, which led to a rise in not only \texttt{Macro} risk, but also \texttt{Supply Chain and Product} and \texttt{Markets and Consumers} risks.

\section{Conclusion}

Risks are ubiquitous to all companies, industries, and society-at-large. Computational modeling of risk factors could better inform analysts, investors, and policymakers. However, how to systematically model risk factors at scale is a challenging question. In this study, we propose a new categorization framework for risks, and further annotate a new dataset over $666$ news articles. We benchmark state-of-the-art NLP models, and analyze a large collection of Bloomberg News articles using the best model. Our analysis demonstrates that modeling risk factors from news could reveal important signals regarding the operations of a company. The aggregated data could further provide information regarding the risks to industries and society.

\section*{Acknowledgements}
This work was conducted while Jiaxin Pei was an intern in Bloomberg's AI Engineering group. We thank Genta Winata, Frederick Zhang, Chuck-Hou Yee, Umut Topkara, and Anju Kambadur for their early feedback on this project.

\bibliography{reference}

\appendix
\section{Risk-Related Lexicon}
\label{sec:appendix-risk-keywords}
In Section~\ref{sec:data}, we discuss a risk-related pre-filtering step to narrow down the dataset for annotation. We rely on a manually curated list of keywords by querying the entire dataset. They are listed below.

\begin{table}[h]
\centering
    \begin{tabular}{|l l l|}
    \hline

    affect & ban & cash \\
    cashflow & challenge & competition \\
    concern & crackdown & cut \\
    debt & decline & decrease \\
    delay & demand & downgrade \\
    drop & fail & finance \\
    harm & hit & impact \\
    inflation & layoff & liable \\
    limit & lose & loss \\
    lowest & operation & plunge \\
    pressure & protest & regulation \\
    restriction & risk & rival \\
    shortage & shrink & slump \\
    strike & struggle & sue \\
    suffer & supply & suspend \\
    tension & unable & uncertain \\
    volatile & warn & weak \\
    worsen & worst &  \\
    \hline
    \end{tabular}
\end{table}

\subsection{Label Count Distribution}
\label{sec:label-count}

\begin{figure}[h]
\centering
\includegraphics[width=55mm]{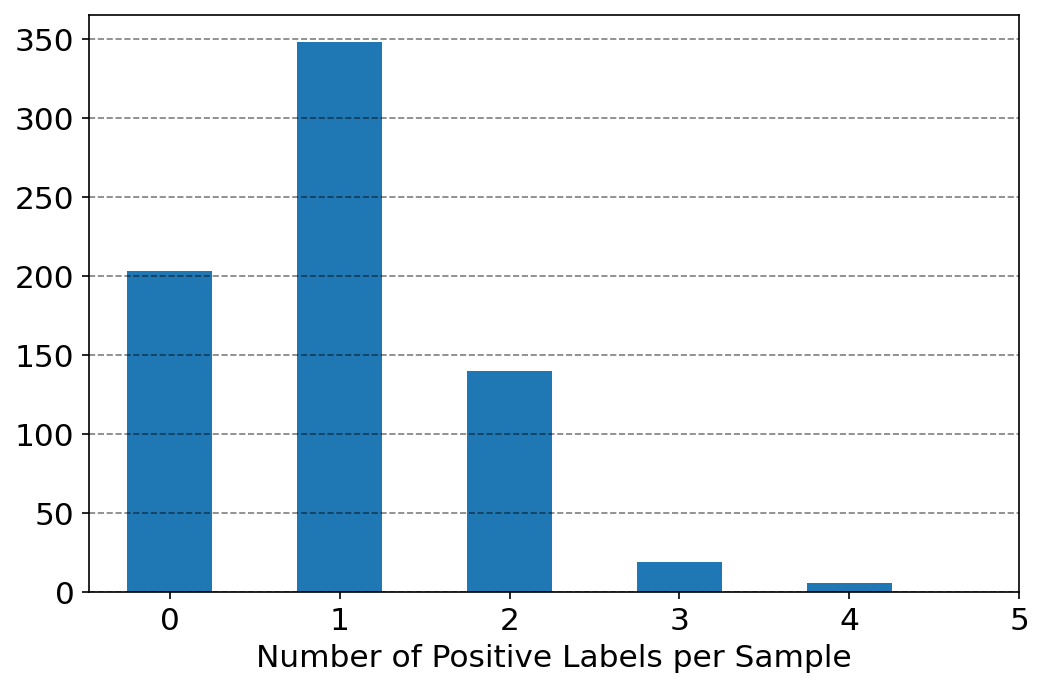}

\caption{
Distribution of number of positive labels per sample. Approximately half of the samples have exactly one risk label.
}
\label{fig:label_count_distribution}
\end{figure}

We consider risk detection as a multi-label classification problem. In Figure~\ref{fig:label_count_distribution}, we show the distribution of positive labels per sample (news article) in our annotated dataset.

\begin{table*}[h]
\centering
\fontsize{9}{11}\selectfont
\begin{tabular}{|p{150mm}|}
\hline
\cellcolor{gray!10} \bf{Tencent's Set for Fastest Growth Since 2018 After Outbreak}  \\
(Bloomberg) -- Tencent Holdings Ltd. picked up millions of new gamers during the global coronavirus outbreak -- yet that surge in mobile play may be slowing as the world's No. 2 economy goes back to work. {[...]} \\
\cellcolor{red!20} Detected risks: \textit{Market} \\

\hline
\cellcolor{gray!10} \bf{Wirecard Shares Surge After Statement on KPMG Audit}  \\
(Bloomberg) -- Wirecard AG , the German payments company trying to move on from reports of alleged questionable accounting methods, said a special investigation has so far found no need to correct financial statements from 2016-2018. {[...]} \\
\cellcolor{red!20} Detected risks: \textit{Management}, \textit{Legal} \\

\hline
\cellcolor{gray!10} \bf{SoftBank Soars After Unveiling \$41 Billion Asset Sale Plan}  \\
(Bloomberg) -- SoftBank Group Corp. surged the most in 11 years after unveiling a plan to raise as much as 4.5 trillion yen (\$41 billion) over the coming year to buy back stock and slash debt, addressing concerns about its exposure to money-losing businesses during the coronavirus pandemic. {[...]} \\
\cellcolor{red!20} Detected risks: \textit{Finance} \\

\hline
\cellcolor{gray!10} \bf{Twitter Surges After Activists Seek to Replace CEO Dorsey}  \\
(Bloomberg) -- Twitter Inc. shares rose in early trading Monday after Bloomberg reported that activist investors have built a sizable stake in the social media company and are pushing for changes, including possibly replacing co-founder and Chief Executive Officer Jack Dorsey. {[...]} \\
\cellcolor{red!20} Detected risks: \textit{Management} \\

\hline
\end{tabular}
\caption{Sample news articles where the sentiment is \textit{Positive} but company risks are detected. Due to space limitation, only the first paragraphs are shown.}
\label{tab:sample-news-sentiment}
\end{table*}

\section{Sample News Articles}

In Table~\ref{tab:sample-news-sentiment}, we show sample articles where the sentiment analysis results are ``Positive,'' while various risk factors are detected.

\end{document}